\documentclass[preprint,11pt,authoryear]{elsarticle}

\usepackage{amssymb}
\usepackage{amsmath}

\usepackage{pgfplots}
\usepackage{graphicx}  

\usetikzlibrary{patterns,pgfplots.groupplots,arrows.meta}
\pgfplotsset{compat=1.18}
\usepgfplotslibrary{polar}

\usepackage{booktabs,tabularx,siunitx,makecell,pifont,xcolor}

\usepackage[hidelinks]{hyperref}

\usepackage{lineno}
\journal{Computer Vision and Image Understanding}
\begin{document}
\begin{frontmatter}

\title{FOOTPASS: A Multi-Modal Multi-Agent Tactical Context Dataset for Play-by-Play Action Spotting in Soccer Broadcast Videos}

\author[inst1,inst2]{Jeremie Ochin\corref{cor1}}
\ead{jeremie.ochin@minesparis.psl.eu}

\author[inst2]{Raphael Chekroun}
\ead{raphael.chekroun@footovision.com}

\author[inst1]{Bogdan Stanciulescu}
\ead{bogdan.stanciulescu@minesparis.psl.eu}

\author[inst1]{Sotiris Manitsaris}
\ead{sotiris.manitsaris@minesparis.psl.eu}

\cortext[cor1]{Corresponding author.}

\affiliation[inst1]{
  organization={Center for Robotics, Mines Paris - PSL},
  addressline={60 Bd Saint-Michel}, 
  city={75006 Paris},
  country={France}
}

\affiliation[inst2]{
  organization={Footovision},
  addressline={17 Rue Saint-Augustin}, 
  city={75002 Paris},
  country={France}
}

\begin{abstract}

\small{Soccer video understanding has motivated the creation of datasets for tasks such as temporal action localization, spatiotemporal action detection (STAD), or multi-object tracking (MOT). The annotation of structured sequences of events (who does what, when, and where) used for soccer analytics requires a holistic approach that integrates both STAD and MOT. However, current action recognition methods remain insufficient for constructing reliable play-by-play data and are typically used to assist rather than fully automate annotation. Parallel research has advanced tactical modeling, trajectory forecasting, and performance analysis, all grounded in game-state and play-by-play data. This motivates leveraging tactical knowledge as a prior to support computer-vision-based predictions, enabling more automated and reliable extraction of play-by-play data.

We introduce \textit{Footovision Play-by-Play Action Spotting in Soccer Dataset} \linebreak (FOOTPASS), the first benchmark for play-by-play action spotting over entire soccer matches in a multi-modal, multi-agent tactical context. It enables the development of methods for player-centric action spotting that exploit both outputs from computer-vision tasks (e.g., tracking, identification) and prior knowledge of soccer, including its tactical regularities over long time horizons, to generate reliable play-by-play data streams. These streams form an essential input for data-driven sports analytics.}

\end{abstract}

\begin{graphicalabstract}
\includegraphics[width=.99\linewidth]{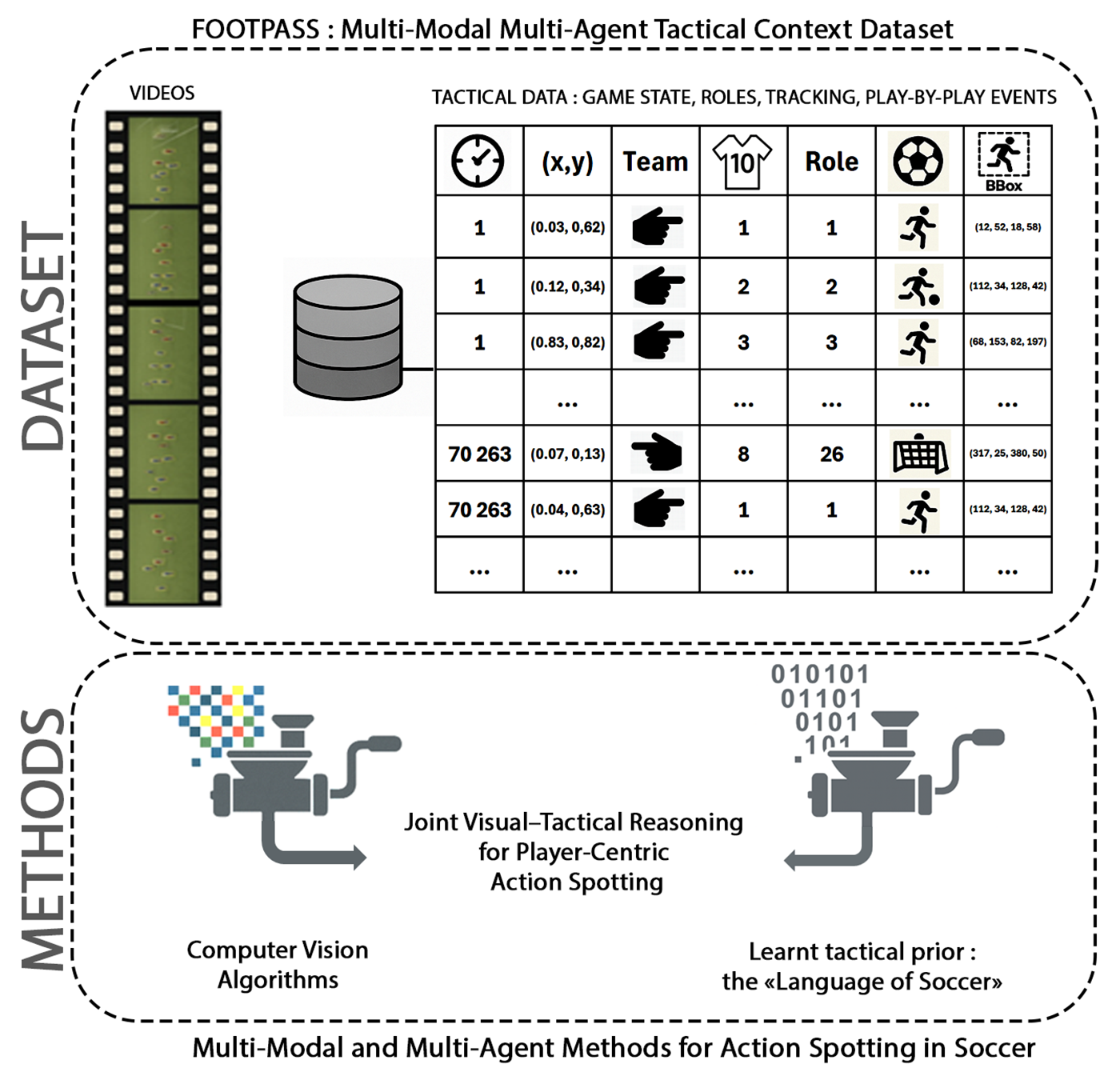}
\end{graphicalabstract}


\end{frontmatter}



\section{Introduction}
\label{Intro}

Soccer stands as the most popular sport in the world, followed by millions of fans and played in over 200 countries. Its cultural significance and global audience have also made it one of the most data-rich sports, as each competition generates hours of video and every match produces millions of positional data points. In recent years, technological progress has transformed how this wealth of information is analyzed, giving rise to what is broadly referred to as soccer analytics, a field that integrates computer vision, machine learning, and domain knowledge to extract meaningful insights from the game \citep{cuevas2020}.

While data-driven methods have long supported broadcasters and fans through visual enhancements and post-match statistics, the most profound shift has occurred within professional clubs themselves. Elite teams increasingly use advanced data analytics to evaluate player performance, optimize training sessions, and model tactical systems. Clubs now rely on video-based analytics not only to study their opponents’ strategies but also to refine their own collective play, analyzing recurring tactical patterns, spatial occupation, and phase transitions across matches. As highlighted by \citet{wangTacticAI2024}, the advent of AI-assisted tactical modeling and forecasting tools, such as TacticAI, has accelerated this transformation by enabling the interpretation of spatiotemporal data at a level of granularity previously achievable only by expert analysts.

This evolution reflects a broader trend: soccer is no longer observed solely through the lens of visual intuition but increasingly through structured, data-centric representations that quantify who does what, when, and where on the pitch. These structured descriptions, commonly known as play-by-play or event data, constitute the foundation of tactical analysis and performance evaluation. Yet, constructing them at scale still requires extensive manual annotation. This motivates research at the intersection of computer vision and soccer analytics, with the goal of automatically generating reliable, player-centric play-by-play data streams from broadcast videos

\subsection{Definitions}
\label{definitions}
\bigskip
In the sports analytics literature, the term \textit{play-by-play data} refers to a structured chronological record of atomic match events, each specifying who performed an action, what the action was, when it occurred, and where on the field it took place \citep{davis2024}. In soccer, two main types of events can be distinguished. \textit{On-ball events} are instances where a player makes contact with the ball: ball drives, passes, crosses, headers, throw-ins, shots, tackles, or blocks. They occur frequently, represent the vast majority of actions in a match, and their sequence captures the flow of play and reveals team tactics. In contrast, \textit{sparse events} are rarer and typically correspond to decisive highlights (e.g., goals), infractions (e.g., fouls, offsides), or referee interventions (e.g., yellow/red cards, VAR checks). Set pieces such as corners or free kicks are usually encoded as passes or shots within on-ball sequences, but are annotated explicitly in sparse event datasets. This work focuses on on-ball play-by-play data.

Play-by-play data thus describe the match as a chain of discrete, player-centric events, enabling the reconstruction of full match dynamics. Such representations form the standard input for tactical analysis, outcome modeling, and performance evaluation. Play-by-play data are used in \textit{soccer} \citep{aalbers2019,korte2019,simpson2022,mendes2024,anzer2025,yeung2025}; \textit{basketball} \citep{vracar2016,damoulaki2025,sun2025mvp}; \textit{handball} \citep{mortelier2024}; and \textit{American football} \citep{otting2021}. This widespread use underscores their centrality in sports analytics.

Alongside play-by-play records, \textit{game-state data} refers to the spatiotemporal localization of players on the pitch together with their identities. This spatiotemporal representation of the game’s dynamics is a standard input for tasks such as tactical analysis and multi-agent trajectory modeling. It is used, for example, in \textit{soccer} \citep{bialkowski2014,martens2021,forcher2022,raabe2022,capallera2024,ogawa2025}; \textit{basketball} \citep{sha2017,felsen2018}; \textit{handball} \citep{mortelier2024}; and \textit{hockey} \citep{lucey2013}. In soccer, play-by-play data can be regarded as a subset of game-state data, filtered to the players currently performing a ball-related action and augmented with an action label.

Given their central role in soccer analytics, both play-by-play and game-state data are essential representations to be extracted through automated video understanding.

\subsection{Bridging Perception and Tactical Reasoning for Action Spotting through Multi-Modal, Multi-Agent Representations}
\label{bridge}
\bigskip

Thanks to dedicated perception datasets, the computer vision subtasks required for soccer game-state reconstruction, such as multi-object tracking, player (re-)identification, jersey number recognition, field localization, and camera calibration, are improving steadily, making game-state recovery increasingly automated. This trend is reflected in the increasing metrics reported for these tasks in the last four editions of the SoccerNet Challenges \citep{giancola2022,cioppa2024a,cioppa2024b,giancola2025}. In contrast, the precise and fine-grained annotation of play-by-play data remains largely a manual, time-consuming process carried out by trained expert operators, who must scrub through the footage to label actions accurately \citep{cartas2022,Bassek2025}.

Current state-of-the-art Spatiotemporal Action Detection (STAD) methods provide a natural starting point for reconstructing play-by-play data from raw soccer broadcast video, as they jointly perform action detection and actor localization. However, although promising, these methods still fall short of the performance required for exhaustive event coverage in soccer analytics, since achieving high recall would produce numerous false positives that must be filtered out by human annotators \citep{singh2023,wang2023,ochin2025gnn}.

Several factors account for these limitations. In particular, the visual conditions of broadcast footage often make reliable detection difficult, as variations in weather, illumination, and shadows combine with occlusions, motion blur, rapid camera movements, frequent visual ambiguities, and replays, to obscure certain actions. Second, existing STAD models often lack contextual understanding, treating detections almost in isolation: they are usually trained on short clips without explicitly reasoning about the tactical intentions of players. This leaves them blind to the structured dependencies that govern soccer. The problem is especially acute in high-recall settings, where many false positives could be avoided by incorporating long-range temporal and game-state contexts \citep{ochin2025}.

Although these approaches may continue to improve, \citet{ochin2025} demonstrated that operating at a more abstract modeling level, centered on players as entities in an evolving system, makes it possible to leverage the \textit{language of soccer}: the tactical and temporal regularities and dynamics that structure the game over time spans much longer than the short clips typically used in conventional video understanding. In this way, even imperfect sequences of action predictions can be refined by exploiting such \textit{tactical priors}.

Therefore, a path toward reliable player-centric action spotting lies in approaches that bridge perception with tactical reasoning. Yet, no existing public dataset provides both: broadcast video annotated at the play-by-play level together with the tactical state information necessary to integrate higher-level reasoning.

To support research at the intersection of computer vision and tactical modeling, we introduce \textit{Footovision Play-by-Play Action Spotting in Soccer Dataset} (FOOTPASS). The dataset provides human-annotated play-by-play data together with multi-modal, multi-agent tactical data aligned with full-length broadcast videos of soccer matches. This includes extended game-state information (player positions and velocities on the pitch, team memberships, jersey numbers, and roles), as well as single player tracking data in screen space.

FOOTPASS is designed to serve both communities: researchers working on fundamental computer vision subtasks and those exploring methods to fuse vision-based data with tactical priors to enhance sequences of spotted events. It establishes a shared foundation for advancing methods that either strengthen individual components of the pipeline or integrate perception with tactical modeling, with evaluation centered on the core benchmark: reliable player-centric play-by-play action spotting.

\subsection{Contributions}
\label{contrib}
\bigskip
This paper makes two main contributions. First, it introduces FOOTPASS, the first dataset for play-by-play action spotting in full-length soccer broadcast videos, designed within a multi-modal and multi-agent tactical context. Second, it benchmarks several existing methods to evaluate their ability to generate precise play-by-play records.

FOOTPASS is thus released as a resource for the community, offering a basis for future research on both improved perception modules and the fusion of vision-based data with tactical priors to deliver reliable play-by-play streams for scalable and data-driven soccer analytics.

\section{Related Work}
\label{relatedwork}
\subsection{Soccer Video Action Recognition Datasets}
\label{soccVD}
\bigskip

This section review the available public datasets for action recognition in Soccer. Their main characteristics are compared in Table \ref{tab:soccer_datasets}.

Early large-scale soccer datasets primarily focused on the detection of \textit{sparse events} such as goals, fouls, cards, and offsides. These annotations captured key highlights of the match but did not provide the dense sequence of on-ball actions that reflects tactical organization and team play.

\newcommand{\TAL}{\ding{51}}          
\newcommand{\STAD}{\(\blacksquare\)}  
\newcommand{\AS}{\(\bullet\)}
\newcommand{\Video}{\(\blacktriangle\)} 
\newcommand{\SpatioTemp}{\(\blacklozenge\)}

\begin{table}[t]
\centering
\small
\sisetup{group-separator={,},group-minimum-digits=4}
{
\renewcommand{\arraystretch}{1.25}

\begin{tabularx}{\textwidth}{l
  >{\centering\arraybackslash}p{1.1cm}
  >{\centering\arraybackslash}p{1.2cm}
  S[table-format=2.0]
  >{\centering\arraybackslash}p{1.5cm}
  >{\raggedright\arraybackslash}X}
\toprule
\makecell[l]{Dataset} & \makecell{Task} & \makecell{Event\\type} & {\#Classes} & \makecell{Modality} & \makecell[l]{\#Events} \\
\midrule
Comprehensive Soccer   & \TAL{} & Sparse & 11 & \Video{} & 6,850 \\
SoccerNet v1           & \AS{} & Sparse &  4 & \Video{} & 6,637 \\
SoccerNet v2           & \AS{} & Sparse & 17 & \Video{} & 110,458 \\
SoccerDB               & \TAL{} & Sparse & 11 & \Video{} & 37,715 \\
SoccerReplay-1988      & \TAL{} & Sparse & 24 & \Video{} & $\sim$150,000 \\
Ball Action Spotting   & \AS{} & On-ball & 12 & \Video{} & 12,433 \\
STAD Multisports       & \STAD{} & On-ball & 15 & \Video{} & 12,254 \\
Integrated Dataset     & \AS{} & On-ball & 45 & \SpatioTemp{} & 11,137 \\
\bottomrule
\end{tabularx}
} 

\caption{Comparison of public soccer datasets across key dimensions. \textbf{Task markers:} \TAL{} = Temporal Action Localization (TAL), \STAD{} = Spatiotemporal Action Detection (STAD), \AS{} = Action Spotting (AS). \textbf{Modality markers:} \Video{} = Video, \SpatioTemp{} = Spatiotemporal data (position on the pitch)}
\label{tab:soccer_datasets}
\end{table}

\citet{yuDataset2018} introduced the Comprehensive Soccer dataset, later extended by \citet{fengSSET2020} to encompass 350 broadcast videos totaling 282 hours. Its annotations include shot boundary detection (types of field of view, replays, and transition types), 11 classes of sparse events with temporal boundaries (start and end frames), and limited tracking annotations covering 40 shots (approximately 13 minutes of play). However, no information is provided about the active players performing the actions.

\citet{giancolaSoccNetv1_2018} released the SoccerNet dataset and proposed the task of action spotting in soccer, where the goal is to locate the anchor time of an event rather than its temporal boundaries (the latter being the focus of Temporal Action Detection or Temporal Action Localization). The original dataset contained 500 complete matches, amounting to 764 hours of video. With SoccerNet v2, published by \citet{deliegeSoccNetv2_2021}, annotations expanded from 4 to 17 sparse event classes, with the addition of shot segmentation annotations and a replay grounding task. Subsequent works \citep{cioppaSoccNetv3_2022, cioppa_SoccNetTrack_2022, Gutierrez-Perez_2025_CVPR} further extended the dataset by incorporating cross-view correspondences between main shots and replays, field line and jersey number annotations, tracking data for all visible players across 200 sequences of 30 seconds, and three-dimensional ball localization annotations on replay sequences.  

Similarly, \citet{yudongSoccDB_2020} introduced the SoccerDB dataset, comprising 343 matches divided into clips ranging from 3 to 30 seconds. The annotations cover 11 classes of sparse events with temporal boundaries, alongside bounding boxes for players and the ball. Nevertheless, no tracking data are included, and event annotations do not identify the active player.

More recently, \citet{rao2025} released the SoccerReplay-1988 dataset, a large-scale multi-modal resource comprising 1,988 broadcast matches from six European leagues (2014/15--2023/24 seasons). Its annotations include 24 classes of sparse events, automatically aligned with around 150,000 broadcast commentaries, as well as rich metadata about matches, teams, coaches, and referees. While the dataset is notable for its scale and the integration of natural language commentaries, it does not provide tracking data, player positions, or game-state annotations. Consequently, it is primarily suited for sparse event spotting and video-language research, rather than dense on-ball STAD, play-by-play action spotting or tactical modeling.

In 2023, a Ball Action Spotting dataset was incorporated into SoccerNet, with 5 complete matches initially annotated with 2 classes of ball-related events (drive and pass). This later evolved into 12 classes of on-ball events and was further extended into a Team Ball Action Spotting task in 2025 \citep{cioppa2024a, cioppa2024b, giancola2025}. Despite these advances, the existing on-ball event annotations do not include the localization or identity of the active players.

To our knowledge, the only public dataset that includes both action classes and active player localization in soccer is the STAD Multisports dataset \citet{liMultiSports_2021}. This dataset spans four sports, basketball, volleyball, soccer, and aerobic gymnastics, and covers 66 action categories with 800 clips per sport, each averaging 750 frames. Actions are annotated through a class label and an action tube, i.e., a sequence of bounding boxes tracking the active player during the temporal extent of the action. For soccer, this amounts to 12,254 action instances across 15 classes and 225,000 bounding boxes. However, it lacks broader tracking data beyond the active player and does not provide player identities.

Finally, the Integrated Dataset of Spatiotemporal and Event Data in Elite Soccer published by \citet{Bassek2025} addresses the scarcity of public resources that combine play-by-play and game-state data. This dataset includes frame-by-frame game-state information for seven Bundesliga matches, together with hierarchically structured discrete events categorized into player, team, and referee actions. It also provides tactical information such as team formations and jersey numbers. Importantly, the positional data represents a true ground truth, captured by an expensive multi-view camera system that tracks all players simultaneously across the full pitch. However, the corresponding broadcast video footage is not publicly available, which limits the possibility of directly synchronizing the annotations with other video sources. Moreover, the authors note that, with only seven matches, the dataset provides a limited sample size for deriving robust conclusions in match analysis, as representative studies typically require much larger datasets.


\subsection{Spatiotemporal Action Detection Methods}
\label{stadMethods}
\bigskip
Spatiotemporal Action Detection (STAD) has received increasing attention in recent years, driven by its broad range of real-world applications in surveillance, sports analysis, and autonomous driving \citep{wang2023}. This growing interest has led to the development of deep learning approaches that can be broadly grouped into two families: frame-level and clip-level methods \citep{liMultiSports_2021}. Frame-level models assign bounding boxes and action labels independently at each frame and subsequently integrate predictions over time. In contrast, clip-level models, often referred to as \textit{action tubelet detectors}, jointly capture temporal context and localize actions across sequences of frames. A more detailed overview of these families is provided in the survey by \citet{wang2023}.

Within clip-level approaches, the Track-Aware Action Detector (TAAD) proposed by \citet{singh2023} generates per-actor, per-frame action predictions by first detecting and tracking actors, then aggregating features along their trajectories with a fine-tuned 3D CNN and ROI Align \citep{maskRCNN2017}, followed by a Temporal Convolutional Network. This design enables TAAD to achieve state-of-the-art performance on public STAD benchmarks, while showing robustness to camera motion.

In contrast, \citet{peral2025} introduced a soccer-specific approach focused on temporally accurate detection of passes and receptions. Their method also begins with tracking potential acting players, but instead of sampling features from the global feature map along the track, it constructs player-centric video tubes and extracts features directly from these cropped sequences to estimate per-frame ball possession.

These approaches are well suited to automated soccer analytics, where reconstructing the game-state typically precedes event annotation, enabling actions to be directly associated with player identities. However, their precision in high-recall settings remains low, primarily because they lack contextual understanding.

\subsection{Multi-Modal and Multi-Agent Action Spotting in Soccer}
\label{multiAgtMultiModMethods}
\bigskip
Recent research has explored moving beyond pixel-based detections toward representations that integrate structured, multi-agent, and multi-modal information, enabling richer contextual reasoning over soccer dynamics. These methods treat the game as an evolving system of interdependent agents, in contrast to conventional spatiotemporal action detection approaches that focus on per-player feature aggregation along motion paths, and therefore capture only limited structured inter-player context.

\citet{ochin2025gnn} proposed a Graph Neural Network (GNN)-based extension of the Track-Aware Action Detector (TAAD) that explicitly incorporates game-state information, such as player positions, velocities, and team memberships, alongside visual features extracted by a 3D CNN. By encoding local inter-player relationships as a spatio-temporal graph, the model captures important contextual information and jointly learns visual and tactical representations, which improve its action-spotting performance. Experiments on a dedicated private dataset demonstrated that fusing visual and structured features leads to substantial precision gains, particularly in the high-recall regimes required for exhaustive event coverage. While this method has the advantage of being trainable end-to-end, it still lacks long-range temporal context, operating on short clips of 50 frames, and has not yet been tested on full-length broadcast matches and public datasets.

To address these temporal limitations, \citet{ochin2025} introduced the \textit{Denoising Sequence Transduction} (DST) model, which extends STAD by integrating additional structured information specific to coordinated multi-agent domains, such as soccer, and by operating on long sequences of actions. The approach uses noisy, context-free, player-centric STAD predictions as a pixel-based prior and then denoises these sequences using role-based structured features, including player positions, velocities, and team context. In this setting, DST produces coherent and tactically plausible action sequences and achieves significant improvements in both precision and recall in high-recall regimes compared to purely visual baselines, while remaining computationally efficient on a single GPU.

Together, these works outline a multi-modal, multi-agent paradigm for action spotting in soccer, bridging the gap between computer vision and tactical modeling. They demonstrate how the integration of game-state reasoning with vision-based detections can lead to more reliable and interpretable play-by-play predictions, an approach directly aligned with the motivation of the FOOTPASS benchmark.

Implementations of both models, retrained on the FOOTPASS dataset, are used as baseline benchmarks, providing reproducible reference points and enabling future comparisons by the research community.

\section{The FOOTPASS Dataset}

At a glance, FOOTPASS provides broadcast video aligned with human-annotated play-by-play data of on-ball events, player identities (via jersey numbers, team memberships, and roles), single-player tracklets, and game-state variables (positions and velocities). The dataset spans 54 full matches and is designed to benchmark reliable player-centric action spotting in a multi-modal, multi-agent tactical context, where the core task is to predict spatiotemporal sequences of actions, \textit{i.e.} identifying \textit{who} performs \textit{what}, \textit{where}, and \textit{when}, directly from broadcast video and auxiliary information.

A distinctive feature of FOOTPASS is the joint availability of broadcast video and tactical context. Ground-truth annotations are provided for play-by-play and game-state data, while tracking data consists of single-player tracklets. Importantly, the play-by-play ground truth always specifies the acting player’s identity (via jersey number), even if no corresponding bounding box is available, since these annotations are provided manually. This setup reflects the natural difficulties of broadcast analysis, where occlusions, replays, and motion blur often obscure parts of the match.

FOOTPASS thus provides the means to bootstrap research at the intersection of perception and tactical modeling, with improvements in perception modules evaluated through their contribution to the core benchmark of reliable play-by-play action spotting.

The remainder of this section details the dataset construction (Section~\ref{subsecDataConst}) and its global statistics (Section~\ref{subsecStats}).

\subsection{Dataset Construction}
\label{subsecDataConst}
\bigskip
The dataset was curated with an emphasis on size, quality, and diversity, enabling researchers to test hypotheses and train deep neural networks under conditions that closely mirror practical applications such as automated play-by-play annotation.

\paragraph{\textbf{Size of the dataset}} We provide data from 54 full-length soccer matches. While this number is smaller than the number of matches in the datasets of \citet{yudongSoccDB_2020} and \citet{deliegeSoccNetv2_2021}, which comprise 347 and 500 matches respectively, it is comparable in terms of the total number of annotated events. Specifically, our dataset contains 102{,}992 events, compared to 37{,}715 and 110{,}458 for the two aforementioned datasets. In addition, our dataset addresses the scarcity of joint spatiotemporal and event data in soccer \citep{Bassek2025}, although the positions of players not visible in the broadcast footage are inferred, as explained in the next section.

\paragraph{\textbf{Quality of the data}} We provide broadcast videos recorded natively in Full HD (1920×1080), at 25 fps, and properly synchronized with the released play-by-play and game-state annotations.

\paragraph{\textbf{Diversity of the data}} The matches were selected from major European leagues and competitions of the season 2023/24 (in alphabetical order): the French Ligue~1, the German Bundesliga, the Italian Serie~A, the Spanish La Liga, and the UEFA Champions League. In total, 50 different teams are represented.

\subsubsection{Game-State Reconstruction}
\label{subsubGSRec}
\bigskip
The game-state reconstruction is performed sequentially: field lines detection and camera calibration, players detection, localization and tracking, and imputation of missing values. The overall goal of this step of the dataset construction process is to determine, for each frame of the broadcast video, \textit{where} is each \textit{player} on the pitch.

\paragraph{\textbf{Field line detection and camera calibration}} The first step of the game-state reconstruction process is to detect specific field lines and landmarks in the video using well-established deep learning approaches to image segmentation \citep{minaeeSurveyImSeg2022}, trained on a private dataset of field lines and landmarks. Given the known dimensions of these field lines and landmarks, a field-to-image homography is then estimated for each frame of the broadcast video, excluding replays. Camera intrinsics (including focal length) and extrinsics are recovered by decomposing the homography under planar constraints \citep{Hartley_Zisserman_2004}, and camera pose and lens distortion parameters are jointly optimized via nonlinear refinement on all visible landmarks and line-straightness constraints.

\paragraph{\textbf{Player detection, localization, and tracking}} Well-established deep learning–based object detection methods are used to detect players \citep{sunSurveyObjDet2024}, trained on a private dataset. At this stage, an initial triage of bounding boxes is performed to retain only single-player detections, since the presence of multiple players in one box can bias its dimensions and negatively affect subsequent localization. In addition, a pitch mask obtained via segmentation is used to eliminate bounding boxes outside the playing field or its immediate surroundings. Similarly to \citet{cartas2022}, it is assumed that a player contacts the ground at the midpoint between the bottom coordinates of the bounding box, which is then projected from the image plane onto the pitch using the camera model. Tracking is carried out in multiple passes using a custom matching algorithm that combines: team membership prediction, recognized jersey numbers and the roster of numbers present in the match, segmentation-based color profiles of different body regions, deep feature similarity, inter-frame bounding boxes intersection-over-union and distances, and statistical priors on player positions and roles. Overall, this production process of positional data is \textit{FIFA EPTS certified}.

\paragraph{\textbf{Imputation of missing values}} In broadcast videos of soccer, on average 9 to 12 players are simultaneously visible on screen. Since this dataset is constructed from single-player tracking data, there are missing values in the positions of players extracted through the process described above. To facilitate the use of the dataset, missing values are imputed and the trajectories of invisible players are inferred using a custom algorithm that interpolates positions between visible timesteps, subject to constraints on speed and acceleration, as well as statistical priors on relative positioning based on player roles.

\paragraph{\textbf{Player role annotation}} Another important type of tactical data provided in this dataset concerns the \textit{player roles}. These roles depend on the team formation adopted during the game and may vary dynamically as players adjust their positioning and responsibilities. Although teams can adopt numerous formations and role configurations depending on their strategy, we reduced the total number of possible roles to 13 in order to facilitate the use of learning algorithms under limited data conditions. Increasing the number of role categories would lead to a higher-dimensional and more sparsely populated feature space, thereby requiring substantially more annotated data for robust training. 

Player roles were determined through a combination of trajectory analysis, clustering, and expert annotation. The defined roles are: \textit{Goalkeeper, Left Back, Left Central Back, Mid Central Back, Right Central Back, Left Midfielder, Right Midfielder, Defensive Midfielder, Attacking Midfielder, Left Winger, Right Winger, Central Forward, and Right Back}.

\subsubsection{Play-by-play Data Annotation}
\label{pbpAnnotation}
\bigskip

\begin{figure}[h!]\centering
\includegraphics[width=.99\linewidth]{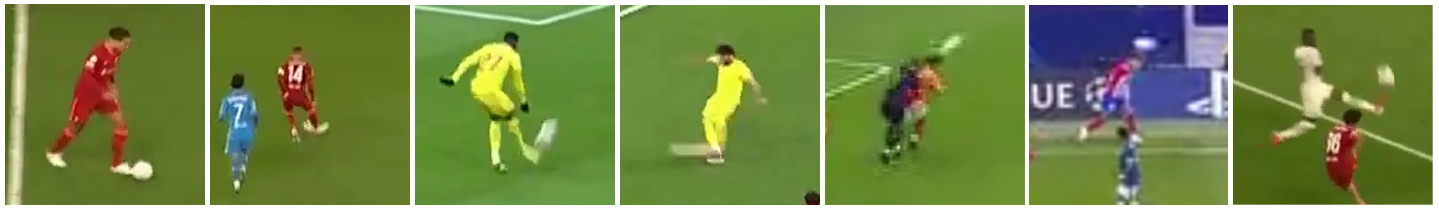}
\caption{Samples of the dataset classes. From left to right : ball-drive, pass, cross, shot, header, throw-in, tackle and ball-block}
\label{fig:classes_samples}
\end{figure}

\paragraph{\textbf{Action classes}}  
The dataset comprises on-ball events, annotated by a professional team using custom-built annotation software designed to streamline the process and ensure both high-quality temporal alignment and accurate identification of the acting player. Events are categorized into eight classes: \textit{drive, pass, cross, shot, header, throw-in, tackle,} and \textit{block}. Samples of those classes can be viewed in Figure \ref{fig:classes_samples}.

\begin{itemize}[{-}]
    \item \textit{Drive:} Occurs when a player receives the ball and maintains possession until passing it to a teammate or losing it to an opponent. The temporal anchor corresponds to the frame in which the player first receives the ball. If the ball is played immediately (without being controlled), a pass is annotated instead, with its anchor defined at the frame in which the ball is struck.
    
    \item \textit{Pass:} Defined when a player strikes the ball with the intent of transferring possession to a teammate. Its temporal anchor is the frame of ball contact.
    
    \item \textit{Cross:} A pass made from outside the penalty area toward a teammate located inside the penalty area, regardless of whether the pass is completed. Its temporal anchor is the frame when the ball is struck.
    
    \item \textit{Shot:} Annotated when the ball is directed toward the goal with the clear intent to score. The temporal anchor is the frame in which the ball is struck.
    
    \item \textit{Header:} Occurs when a player intentionally strikes the ball with the head. The temporal anchor is the frame of head–ball contact.
    
    \item \textit{Throw-in:} A manual throw performed by a player after the ball has completely crossed the touchline. Its temporal anchor is the frame when the ball leaves the player’s hands.
    
    \item \textit{Tackle:} Annotated when a player legally dispossesses an opponent of the ball. The temporal anchor is the frame in which ball contact occurs.
    
    \item \textit{Block:} Occurs when a player intercepts an opponent’s pass or shot. The temporal anchor is the frame when the ball is first touched by the intercepting player.
\end{itemize}

\paragraph{\textbf{Temporal grounding of events}} Given the instantaneous nature of on-ball events—ball touches—the spotting framework introduced by \citet{giancolaSoccNetv1_2018} is adopted. Each event is temporally grounded by a single anchor frame, defined as the frame in which the ball touch occurs. Consequently, events are treated as temporally instantaneous and have no duration.

\paragraph{\textbf{Event representation}} Each event is represented as a tuple \textit{(frame, class, team, jersey)}, where:
\begin{itemize}[{-}]
    \item \textit{frame} is the index of the video frame in which the event occurs (indices start at 0),
    \item \textit{class} is an integer corresponding to the event class (from 0 to 9),
    \item \textit{team} is an integer indicating the team of the acting player (0 for the left team, 1 for the right team), and
    \item \textit{jersey} is an integer corresponding to the jersey number of the acting player.
\end{itemize}

The play-by-play annotations are integrated with the game-state data by augmenting each player’s state with class labels, including a background class for non-action frames.

\subsection{Dataset Statistics}
\label{subsecStats}
\begin{figure}[t]
\centering
\begin{tikzpicture}
\begin{axis}[
    ybar,
    bar width=0.6cm,
    width=13cm,
    height=8cm,
    ylabel={Number of instances ($\times 10^4$)},
    xlabel={Classes of events},
    ymin=0,
    ymax=65000,
    xtick=data,
    xticklabels={pass, drive, header, cross, throw-in, block, shot, tackle},
    xticklabel style={rotate=25, anchor=north},
    ymajorgrids=true,
    grid style=dashed,
    nodes near coords,
    point meta=explicit symbolic,
    every node near coord/.append style={
        font=\small, rotate=90, anchor=west, yshift=2pt
    }]

\addplot[
    ybar,
    fill=gray!50,
    draw=black
] coordinates {
    (1,51424) [49.9\%]
    (2,40183) [39.0\%]
    (3, 4074) [4.0\%]
    (4, 2404) [2.3\%]
    (5, 1926) [1.9\%]
    (6, 1415) [1.4\%]
    (7, 1231) [1.2\%]
    (8,  335) [0.3\%]
};

\end{axis}
\end{tikzpicture}
\caption{Distribution of annotated events across the 8 on-ball classes. Passes and drives dominate, while decisive actions such as shots and tackles are comparatively rare.}
\label{fig:class_distribution}
\end{figure}
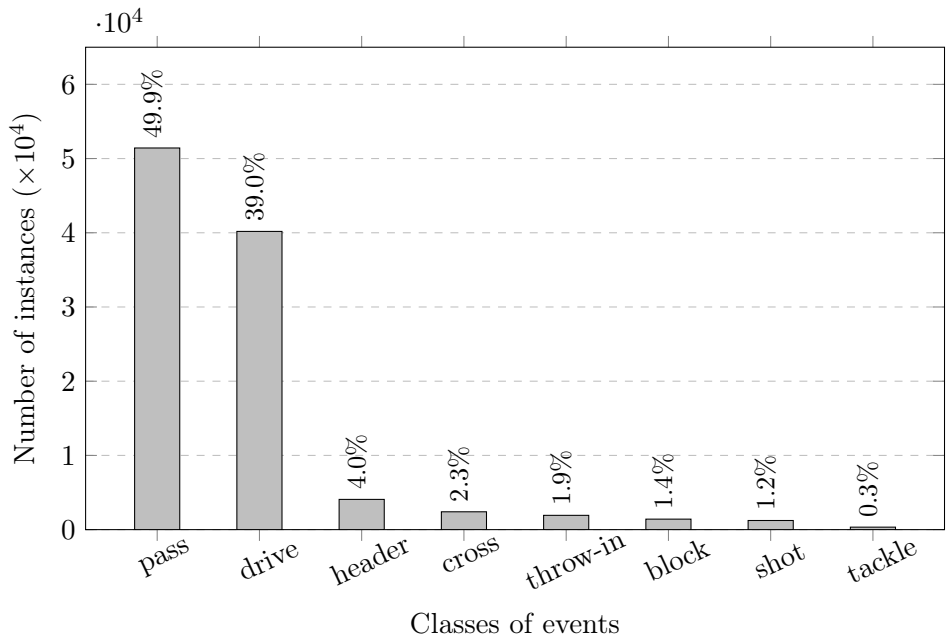

\bigskip
\paragraph{\textbf{Class distribution}} The dataset exhibits a strong class imbalance (Fig.~\ref{fig:class_distribution}). The vast majority of events are \textit{passes} (49.9\%) and \textit{drives} (39.0\%), together accounting for nearly 90\% of all annotations. Less frequent actions include \textit{headers} (4\%), \textit{crosses} (2.3\%), \textit{throw-ins} (1.9\%), \textit{blocks} (1.4\%), \textit{shots} (1.2\%), and \textit{tackles} (0.3\%). This distribution mirrors the natural statistics of soccer, where passes and carries dominate ball interactions, while decisive actions such as shots and tackles are comparatively rare.

\begin{figure}[t]
\centering
\begin{tikzpicture}
\begin{axis}[
    ybar stacked,
    bar width=0.6cm,
    width=12cm,
    height=6cm,
    ymin=0, ymax=100,
    xlabel={Classes of events},
    ylabel={Proportion within class (\%)},
    xtick=data,
    xticklabels={pass, drive, header, cross, throw-in, block, shot, tackle},
    xticklabel style={rotate=45, anchor=east},
    ymajorgrids=true,
    grid style=dashed,
    ytick={0,20,40,60,80,100},
    yticklabel=\pgfmathprintnumber{\tick}\,\%,
    legend cell align=left,
    legend pos=south east
]
\addplot+[draw=black, fill=gray!60] coordinates {
    (1,81.52) (2,85.68) (3,66.74) (4,70.67)
    (5,43.77) (6,78.37) (7,81.56) (8,62.09)
};
\addplot+[draw=black, pattern=north east lines, pattern color=black] coordinates {
    (1,18.48) (2,14.32) (3,33.26) (4,29.33)
    (5,56.23) (6,21.63) (7,18.44) (8,37.91)
};
\legend{BBox available, No BBox}
\end{axis}
\end{tikzpicture}
\caption{Proportion of events per class with bounding-box annotation after interpolation and extrapolation. Coverage is highest for drives, passes, and shots, while headers, tackles, and throw-ins are more affected by occlusion and broadcast editing practices.}
\label{fig:bbox_availability}
\end{figure}
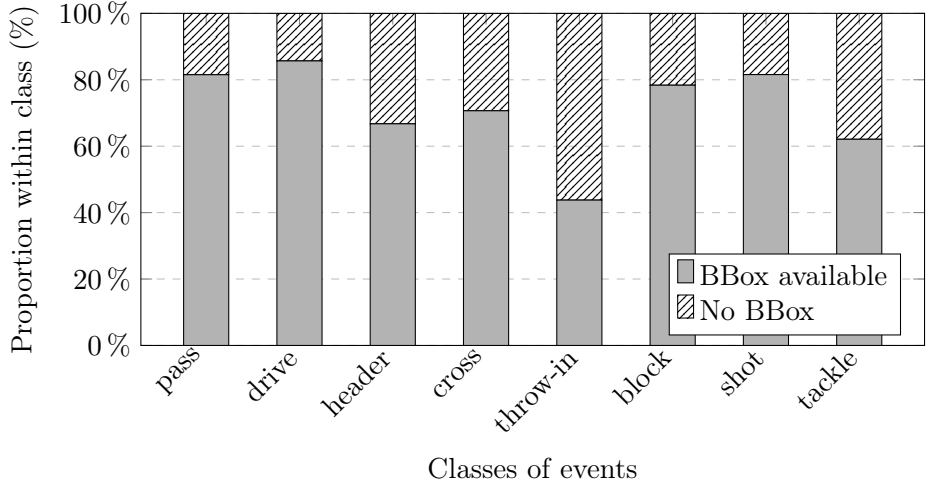

\paragraph{\textbf{Bounding-box coverage}} STAD approaches either predict bounding boxes for actors during inference or require bounding boxes as input to perform action detection on pre-identified players. To support the latter case, FOOTPASS provides single-player tracking bounding boxes. An event is considered covered if the acting player has an associated bounding box after tracklet interpolation to fill gaps shorter than 50 frames (Fig.~\ref{fig:bbox_availability}). On average, 81.5\% of events are covered. Certain classes, such as \textit{blocks} (78.4\%), \textit{crosses} (70.7\%), \textit{headers} (66.7\%), \textit{tackles} (62.1\%), and \textit{throw-ins} (43.8\%), show lower coverage due to their frequent occurrence in crowded areas (e.g., close to penalty boxes), their involvement in duels or player contacts that lead to occlusion, or broadcast practices that cut to replays. By contrast \textit{drives} (85.7\%), \textit{shots} (81.6\%) and \textit{passes} (81.5\%) achieve higher coverage rates, reflecting their greater observability in broadcast footage.

\begin{figure}[t]
\centering
\begin{tikzpicture}
\begin{axis}[
    ybar stacked,
    bar width=0.6cm,
    width=12cm,
    height=6cm,
    ymin=0, ymax=100,
    xlabel={Classes of events},
    ylabel={Proportion within class (\%)},
    xtick=data,
    xticklabels={pass, drive, header, cross, throw-in, block, shot, tackle},
    xticklabel style={rotate=45, anchor=east},
    ymajorgrids=true,
    grid style=dashed,
    ytick={0,20,40,60,80,100},
    yticklabel=\pgfmathprintnumber{\tick}\,\%,
    legend cell align=left,
    legend pos=north east
]
\addplot+[draw=black, fill=gray!60] coordinates {
    (1, 9.79) (2, 8.06) (3, 4.59) (4,10.07)
    (5,46.83) (6, 4.95) (7, 0.49) (8, 6.57)
};
\addplot+[draw=black, pattern=north east lines, pattern color=black] coordinates {
    (1,90.21) (2,91.94) (3,95.41) (4,89.93)
    (5,53.17) (6,95.05) (7,99.51) (8,93.43)
};
\legend{Replay, Live}
\end{axis}
\end{tikzpicture}
\caption{Distribution of events by broadcast mode (live vs.\ replay). Throw-ins frequently fall into replay segments because directors cut away to show the preceding sequence after the ball goes out of play, returning only after the throw-in has been executed. In contrast, headers, blocks, and shots occur almost exclusively during live play (>95\% of the time).}
\label{fig:replay_distribution}
\end{figure}
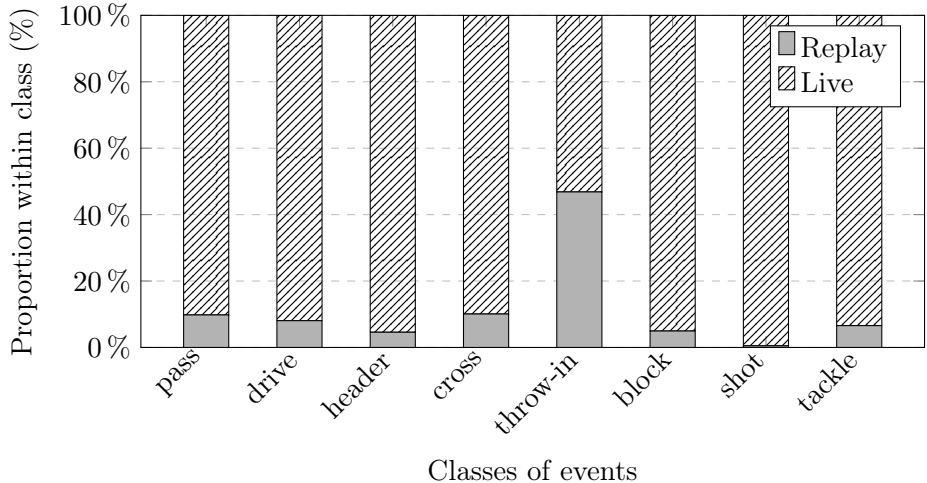

\paragraph{\textbf{Replay vs. live broadcast}} Fig.~\ref{fig:replay_distribution} reports the proportion of events whose anchor frame occurs during live broadcast segments versus replay segments. As expected, most events occur during live play. However, 46.8\% of \textit{throw-ins} fall into replay segments because broadcast directors typically cut to replays of the preceding action after the ball goes out of play and return to live coverage only once the throw-in has already been executed. By contrast, actions such as \textit{headers} (95.4\% live), \textit{blocks} (95\% live), and especially \textit{shots} (99.5\% live) almost always occur during live play. These editing practices disrupts temporal continuity, complicating temporal localization.

\bigskip

\paragraph{\textbf{Summary}} In summary, the dataset provides a realistic distribution of soccer events characterized by strong class imbalance, varying levels of visual observability, and broadcast-specific artifacts such as replays. These properties capture the inherent complexity of real-world soccer video and underscore the importance of approaches that combine low-level perception with contextual reasoning for reliable play-by-play reconstruction.

\section{Experiments}
\label{expAndAnalysis}
\subsection{The FOOTPASS benchmark}
\label{subBenchmark}
\bigskip

\paragraph{\textbf{Benchmark setup}} The 54 matches in FOOTPASS were manually divided into training, validation, and test sets. The validation and test sets were selected to ensure that even the least-represented classes contain at least 15 visible instances in each split. All videos were downsampled from Full HD (1920$\times$1080) to a resolution of 352$\times$640. The resulting split comprises 91{,}327 training instances from 48 matches, 6{,}070 validation instances from 3 matches (indices 18, 24, and 47), and 5{,}595 test instances from 3 matches (indices 0, 36, and 50).

\paragraph{\textbf{Metrics}} Following \citet{ochin2025}, we report overall and per-class Precision and Recall at a low threshold $\tau$ = 0.15, which is slightly above the score expected from a uniform distribution given the number of classes. This configuration prioritizes high recall to minimize missed detections, while aiming to maximize precision within that regime. It is suitable in the context of assisted annotation, where it is faster to discard a false positive than to scrub through the video to find a missed false negative. Evaluation is performed by matching predicted and ground-truth actions of the same player and class within a fixed temporal tolerance of $\pm \delta$ frames around the annotation, with $\delta = 12$. A match is counted as a true positive (TP), unmatched predictions as false positives (FP), and unmatched ground-truth events as false negatives (FN). This procedure enables the computation of Precision and Recall both overall and per class. Additionally, we report per-class Average Precision (AP), computed using the 11-point interpolation method proposed in \citet{pascalVOC2010}. All metrics are reported on the test set.

\subsection{Benchmarked Methods}
\label{subMethods}
\bigskip
To evaluate the benefit of reasoning at the game level using tactical information, three methods are benchmarked on FOOTPASS: TAAD \citep{singh2023}, a purely visual STAD approach that produces per-player predictions; TAAD+GNN, a graph-based extension of TAAD derived from \citet{ochin2025gnn}; and TAAD+DST, a post-processing method applied to TAAD, proposed by \citet{ochin2025}. The latter two methods exploit the multi-modal, multi-agent tactical context provided by FOOTPASS. 

In addition, an ablation study is conducted on the DST model, extending that of the original paper, to demonstrate that the observed improvements stem from reasoning at the game level rather than from learning to process TAAD’s raw predictions more efficiently than standard Non-Maximum Suppression (NMS) and thresholding.

\paragraph{\textbf{Implementation Details – TAAD}} A lightweight implementation of \linebreak TAAD was trained on randomly selected short clips of 50 frames sampled at 25 frames per second (2 seconds of video), each containing at least one annotated action for which the acting player had a bounding box. To mitigate class imbalance during training, no more than 500 samples per class were selected in a given epoch, thereby reducing the ratio between the most and least represented classes. Moreover, since TAAD produces player-centric predictions and up to 22 players can be visible on the field during an action, 21 of whom are background, only 4 to 5 background tracklets were provided as negative samples to the network along with the tracklet containing the action.

Training used the AdamW optimizer \citep{adamW2019} with a learning rate of $10^{-3}$ for the classification head and $5 \times 10^{-5}$ for the pre-trained X3D-L video backbone \citep{feichtenhofer_x3d_2020}, a weight decay of $10^{-4}$ on non-bias parameters, and a batch size of 6. The network was trained for 25 epochs on a single NVIDIA RTX A6000 GPU, with learning rates reduced by a factor of 10 at epochs 10 and 20. Data augmentation was applied to both videos and bounding boxes, including random scaling, rotation, translation, horizontal flipping, and color jittering, using the Albumentations library \citep{albumentations_2020}. Following the recommendations of \citet{hongE2E2022} for handling a dominant background class in action spotting, a label dilation of $\pm1$ frame was applied during training, and the cross-entropy loss weights of the foreground classes were boosted relative to the background.

For evaluation, the trained TAAD network was applied to the full matches of the test set, split into overlapping windows of 50 frames with a stride of 25 frames. Overlapping logits were averaged and stored for further processing. Per-role, per-frame predictions were obtained by applying a softmax and selecting the action with the highest score. As a post-processing step, temporal NMS with a 25-frame window was applied per player and per action class to handle successive actions occurring in quick succession. The resulting predictions were concatenated to produce play-by-play records and matched with the ground-truth annotations.

\paragraph{\textbf{Implementation Details – TAAD+GNN}} A variant of TAAD+GNN based on \citet{ochin2025gnn} was implemented. The same pre-trained X3D-L video backbone was used, together with Edge Convolution layers employing an asymmetric edge function \citep{wang2019EdgeConv}. To incorporate temporal modeling, Temporal Convolutional layers were inserted between Edge Convolution layers instead of creating explicit temporal edges between nodes representing the same player across adjacent time steps. Additionally, a Gated Recurrent Unit \citep{choGRU2014} layer was added before the final classification head.

The optimizer, training parameters, and data augmentation scheme were identical to those used for TAAD, with the addition that position and velocity data were also flipped when the corresponding video frame was flipped horizontally. A focal loss \citep{linFocalLoss2020} was employed to mitigate the dominance of the background class among the 22 player instances. The same evaluation and post-processing procedures as for TAAD were applied.

\paragraph{\textbf{Implementation Details – TAAD+DST}} The DST model was trained using the averaged TAAD logits from the training set and the role-based structured game-state representation described in \citet{ochin2025}. Training was performed on randomly sampled sequences of 750 frames, using the AdamW optimizer with a learning rate of $2.5 \times 10^{-4}$, a weight decay of $10^{-4}$ on non-bias parameters, and a batch size of 96. A single RTX A6000 GPU was used. The network was trained for 10 epochs, with the learning rate reduced by a factor of 10 at epochs 3, 6, and 8. For each epoch, 2000 sequences of 750 frames were randomly sampled from each of the 48 full-length matches in the training set. To mitigate overfitting, data augmentation was performed by randomly mirroring players along the X-axis of the pitch, the Y-axis, or both. Team memberships and roles were adjusted accordingly (e.g., a left winger from team ``right'' becomes a right winger from team ``left'' when mirroring along the X-axis). Since the averaged logits are ordered by role on the pitch, they were reshuffled to reflect the applied symmetries.

For evaluation, the trained DST network was applied to the full matches of the test set, split into contiguous windows of 750 frames. No post-processing was applied, except for thresholding the confidence scores of the predictions. The resulting predictions were concatenated to produce play-by-play predictions and matched with the ground-truth annotations.

\paragraph{\textbf{Implementation Details – TAAD+DST Ablation}} To rule out the possibility that the DST merely learns to post-process the continuous per-player signals produced by TAAD (the logits) into event sequences, the same training procedure as above was applied to \textit{unstructured} data. In this setting, the DST was trained on batches of individual player sequences of logits, together with corresponding tactical features (player position, velocity, and role, the latter provided as a one-hot encoded vector). These features were encoded and decoded by the DST, enabling it to learn how to transform each player’s sequence of logits and tactical data into an event sequence.

While the original DST is trained using three separate cross-entropy losses (one for the frame, one for the player role, and one for the action), this ablated version employs only two: one for the frame and one for the action. The optimizer, training parameters, and data augmentation scheme were identical to those used for TAAD+DST, and the same evaluation procedure was applied.

\subsection{Results and discussion}
\label{resultsAndDiscussion}
\bigskip
\paragraph{\textbf{Overall performance}} At a low confidence threshold suited for assisted annotation ($\tau{=}0.15$), the purely visual TAAD baseline attains high recall but low precision, generating many spurious detections (Table~\ref{tab:PrecRecComp}). Incorporating spatiotemporal inter-player relationship modeling (TAAD+GNN) improves both recall (+3 points) and precision (+19 points), lifting F1 from 35.9\% to 52.1\%. Finally, the model that reasons over long temporal horizons and at the game level (TAAD+DST) achieves the best overall performance, with 68\% precision and 67\% recall, almost doubling the F1 score relative to TAAD (67.5\% vs. 35.9\%). These results indicate that modeling tactical structure and temporal context provides a promising and effective direction for reliable play-by-play extraction in soccer analytics.

\bigskip

\begin{table}[h!]
\centering
\resizebox{\textwidth}{!}{
\begin{tabular}{l|ccc||ccc||ccc||ccc}
\hline
\textbf{Class} & \multicolumn{3}{c||}{\textbf{TAAD}} & \multicolumn{3}{c||}{\textbf{TAAD+GNN}} & \multicolumn{3}{c||}{\textbf{TAAD+DST}} & \multicolumn{3}{c}{\textbf{TAAD+DST (Abl.)}} \\ 
 & \multicolumn{3}{c||}{\small{\citep{singh2023}}} & \multicolumn{3}{c||}{\small{\citep{ochin2025gnn}}} & \multicolumn{3}{c||}{\small{\citep{ochin2025}}} & \multicolumn{3}{c}{\small{\citep{ochin2025}}} \\ 
\cline{2-13}
 & \textbf{PR} & \textbf{REC} & \textbf{F1} & \textbf{PR} & \textbf{REC} & \textbf{F1} & \textbf{PR} & \textbf{REC} & \textbf{F1} & \textbf{PR} & \textbf{REC} & \textbf{F1} \\ \hline
Pass       & 43.8 & 61.0 & 51.0 & 56.7 & 65.1 & 60.6 & \textbf{72.6} & \textbf{72.1} & \textbf{72.3} & 43.9 & 57.9 & 49.9 \\ 
Drive      & 23.9 & 60.0 & 34.2 & 51.6 & 64.3 & 57.3 & \textbf{68.2} & \textbf{68.9} & \textbf{68.5} & 52.1 & 55.1 & 53.6 \\ 
Header     & 10.6 & \textbf{54.1} & 17.7 & 15.7 & 45.9 & 23.4 & \textbf{30.7} & 26.3 & \textbf{28.3} & 24.8 & 38.5 & 30.2 \\ 
Cross      & 16.6 & \textbf{66.4} & 26.6 & 32.0 & 59.1 & 41.5 & \textbf{62.8} & 51.8 & \textbf{56.8} & 47.7 & 38.7 & 42.7 \\ 
Throw-in   & 23.8 & 44.4 & 31.0 & 42.9 & 39.4 & 41.1 & \textbf{65.9} & \textbf{58.6} & \textbf{62.0} & 40.0 & 30.3 & 34.5 \\ 
Block      & 3.2  & \textbf{47.1} & 6.0  & 4.9  & 33.8 & 8.6  & \textbf{23.3} & 14.7 & \textbf{18.0} & 10.8 & 14.7 & 12.5 \\ 
Shot       & 12.3 & \textbf{74.6} & 21.1 & 28.3 & 61.2 & 38.7 & \textbf{63.4} & 67.2 & \textbf{65.2} & 36.6 & 50.7 & 42.5 \\ 
Tackle     & 1.5  & 16.7 & 2.8  & \textbf{2.4}  & \textbf{25.0} & \textbf{4.4}  & 0.0  & 0.0  & 0.0  & 0.0  & 0.0  & 0.0 \\ \hline
\textbf{Overall} & 25.6 & 59.9 & 35.9 & 44.5 & 62.7 & 52.1 & \textbf{68.2} & \textbf{66.8} & \textbf{67.5} & 45.0 & 54.0 & 49.1 \\ \hline
\end{tabular}}
\caption{Comparison of the Precision (\textbf{PR}), Recall (\textbf{REC}), and F1-score (\textbf{F1}) across benchmarked methods at a confidence threshold of 15\% and $\delta$ = 12 frames. Metrics are $\times 10^2$. The best performance per class is highlighted in bold.}
\label{tab:PrecRecComp}
\end{table}

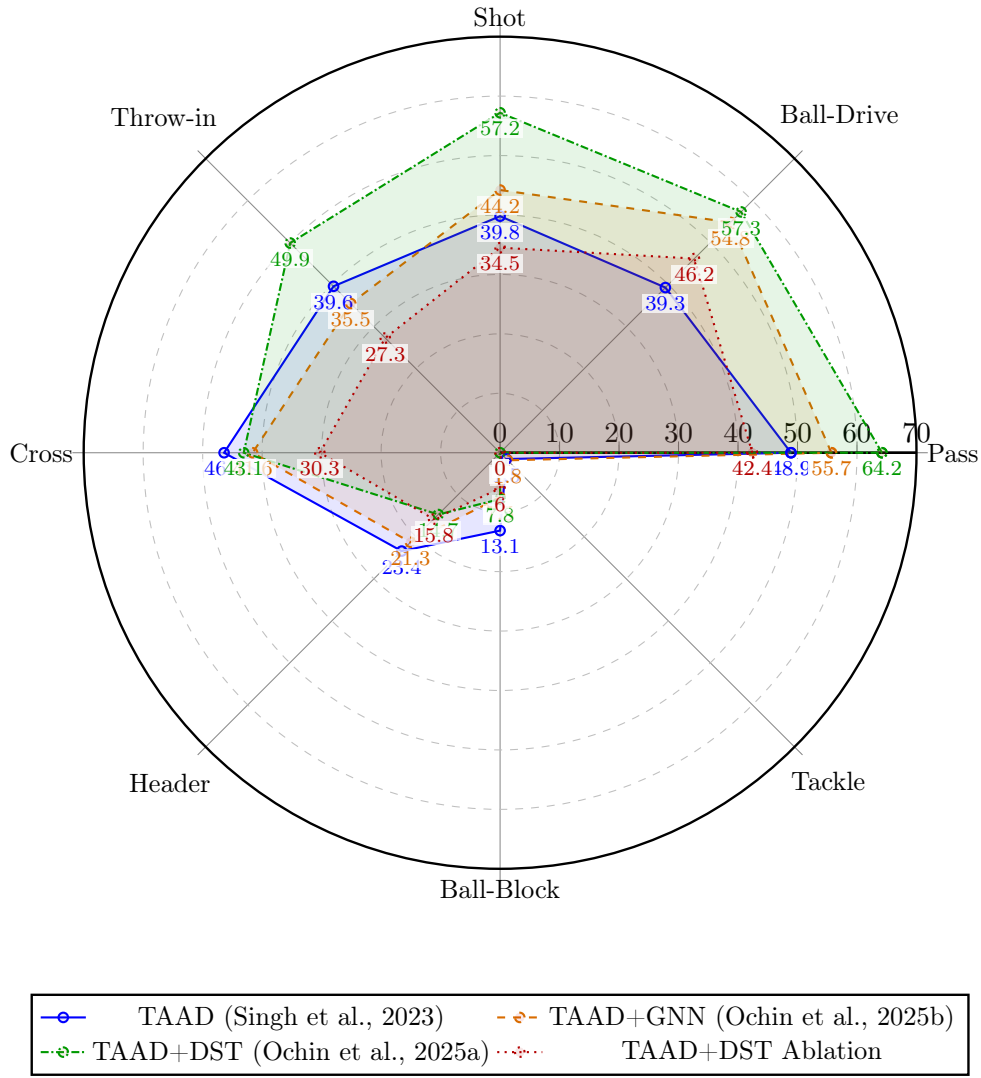
\begin{figure}[h!]
\centering
\begin{tikzpicture}
\begin{polaraxis}[
    width=\textwidth,
    height=\textwidth,
    ymin=0, ymax=70,
    ytick={0,10,20,30,40,50,60,70},
    xmajorgrids, ymajorgrids, grid=both,
    x grid style={black!45}, y grid style={black!25, dashed},
    xtick={0,45,90,135,180,225,270,315},
    xticklabels={Pass,Ball-Drive,Shot,Throw-in,Cross,Header,Ball-Block,Tackle},
    xticklabel style={font=\small, align=center},
    legend style={font=\small, at={(0.5,-0.15)}, anchor=north, legend columns=2},
    line width=0.9pt,
    mark size=1.8pt,
    clip=false,
]

\pgfplotsset{
  mynodes/.style={
    nodes near coords,
    point meta=rawy,
    nodes near coords={\pgfmathprintnumber[fixed,precision=1]{\pgfplotspointmeta}},
    every node near coord/.append style={
      font=\scriptsize,
      fill=white, fill opacity=0.75,
      inner sep=1pt, text opacity=1,
      yshift=-10pt
    },
  }
}

\addplot+[thick, mark=*, color=blue, fill=blue, fill opacity=0.10, mynodes]
coordinates {(0,48.9) (45,39.3) (90,39.8) (135,39.6) (180,46.4) (225,23.4) (270,13.1) (315,1.5) (0,48.9)};
\addlegendentry{TAAD~\citep{singh2023}}

\addplot+[thick, mark=*, dashed, color=orange!85!black, fill=orange!85!black, fill opacity=0.10, mynodes]
coordinates {(0,55.7) (45,54.8) (90,44.2) (135,35.5) (180,41.6) (225,21.3) (270,7.5) (315,1.8) (0,55.7)};
\addlegendentry{TAAD+GNN~\citep{ochin2025gnn}}

\addplot+[thick, mark=*, densely dashdotted, color=green!60!black, fill=green!60!black, fill opacity=0.10, mynodes]
coordinates {(0,64.2) (45,57.3) (90,57.2) (135,49.9) (180,43.1) (225,14.7) (270,7.8) (315,0.0) (0,64.2)};
\addlegendentry{TAAD+DST~\citep{ochin2025}}

\addplot+[thick, mark=*, dotted, color=red!70!black, fill=red!70!black, fill opacity=0.10, mynodes]
coordinates {(0,42.4) (45,46.2) (90,34.5) (135,27.3) (180,30.3) (225,15.8) (270,6.0) (315,0.0) (0,42.4)};
\addlegendentry{TAAD+DST Ablation}

\end{polaraxis}
\end{tikzpicture}
\caption{Comparison of Average Precision (AP, $\times 10^2$) per action class across benchmarked methods, with $\delta$ = 12 frames.}
\label{fig:ap_radar_sorted}
\end{figure}

\paragraph{\textbf{Class-specific behavior}} Performance gains are not uniform across classes (Fig.~\ref{fig:ap_radar_sorted}, Table~\ref{tab:PrecRecComp}). \textit{Drive} and \textit{Pass} show the largest recall improvements, reflecting how both actions depend on player trajectories and inter-player interactions, patterns best captured by TAAD+GNN and TAAD+DST.

The \textit{drive} class is particularly difficult because its annotation depends explicitly on temporal context. The action begins at the first ball control after reception, and without memory, every contact could be misinterpreted as a new drive. Models lacking sequence-level reasoning may over-segment these actions, while TAAD+DST’s temporal modeling correctly anchors the initial touch, explaining its superior performance for this class.

Sparse and spatially constrained events such as \textit{Cross}, \textit{Throw-in} and \textit{Block} show the largest relative precision improvements. These actions often occur near the sidelines or in crowded areas, where context about player positions and team organization may help reduce false detections. \textit{Shot} also benefits strongly (F1 increasing from 21.1\% to 65.2\%), likely reflecting its higher contextual predictability and visibility compared with other action types.

Interestingly, \textit{Header} shows the highest AP with the TAAD baseline, even though TAAD+DST achieves better F1 at low confidence thresholds. This indicates that while DST enhances recall, it slightly degrades the high-precision region of TAAD’s signal for this class. A likely explanation lies in the nature of headers, which are often less controlled or tactically structured than passes or drives. The ball may deflect unpredictably or head in non-strategic directions, making long-range tactical reasoning less effective and sometimes mildly detrimental when denoising. A similar observation applies to \textit{Block}, where the action outcome often reflects reaction rather than intention.

Finally, the \textit{tackle} class remains challenging across methods given its scarcity and frequent occlusions. TAAD+DST did not register true positives for \textit{tackle}, suggesting insufficiently discriminative cues in current features and the need to adjust sequence sampling during training to mitigate class imbalance.

\paragraph{\textbf{Comparison with previous experiments}} Relative to the results reported in prior work, our numbers are lower in several classes, a discrepancy largely explained by dataset scale. The original TAAD+GNN from \citet{ochin2025gnn} was trained on a curated clips dataset with at least 2{,}500 samples per class and evaluated under a temporal action localization protocol, where events have a start and an end, rather than action spotting as in FOOTPASS. Reported results reached a mean Average Precision (mAP) of 49.7\% at a temporal IoU threshold of 0.5, compared with 32.8\% mAP with $\delta$ = 12 frames for TAAD+GNN and 36.8\% mAP for TAAD+DST, both trained on FOOTPASS. Although not directly comparable, these differences align with the stronger supervision available in the former dataset and its controlled clip-level context, which excluded full broadcast artifacts such as replays, occlusions, and long off-ball sequences. The sharper performance drop observed here is therefore consistent with the increased difficulty of broadcast-level data and the stronger class imbalance in FOOTPASS.

The original TAAD from \citet{ochin2025} was trained on the same rich clips dataset as above, and the DST was trained on a much larger set of full matches (240 vs.\ 48 in FOOTPASS), providing both higher-quality prior detections and greater temporal diversity for sequence learning. Reported results on that private dataset reached 78.7\% overall precision and 75.8\% overall recall, higher than those observed on FOOTPASS, respectively 68.2\% and 66.8\%, confirming that the model performances scales with data quantity.

Despite using roughly four times less data than the private dataset of \citet{ochin2025}, these experiments demonstrate that with appropriate data augmentation, such data-hungry methods can still be trained effectively and reproduce their relative improvements over visual baselines. This highlights the value of FOOTPASS as a public benchmark for studying how multi-modal, tactically grounded reasoning can be learned under realistic, resource-limited conditions.

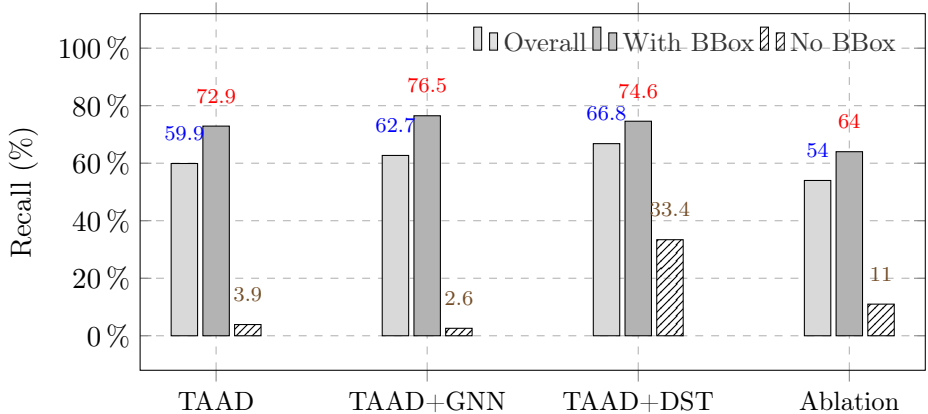
\begin{figure}[h!]
\centering
\begin{tikzpicture}
\begin{axis}[
  width=0.95\textwidth,
  height=0.5\textwidth,
  ybar,
  bar width=10pt,
  enlargelimits=0.12,
  ymin=0, ymax=100,
  ytick={0,20,40,60,80,100},
  yticklabel=\pgfmathprintnumber{\tick}\,\%,
  ylabel={Recall (\%)},
  symbolic x coords={TAAD,TAAD+GNN,TAAD+DST,Ablation},
  xtick=data,
  xticklabel style={font=\small},
  legend style={
    at={(0.98,0.98)},
    anchor=north east, 
    draw=none,
    fill=none, 
    fill opacity=0.8,  
    font=\small,
    legend columns=-1 
},
  grid=both, ymajorgrids,
  grid style={dashed,black!30},
  nodes near coords,
  nodes near coords align={vertical},
  every node near coord/.append style={font=\scriptsize, yshift=5pt}
]

\addplot+[fill=black!15, draw=black] coordinates {
  (TAAD,59.9) (TAAD+GNN,62.7) (TAAD+DST,66.8) (Ablation,54.0)
};
\addlegendentry{Overall}

\addplot+[fill=gray!60, draw=black] coordinates {
  (TAAD,72.9) (TAAD+GNN,76.5) (TAAD+DST,74.6) (Ablation,64.0)
};
\addlegendentry{With BBox}

\addplot+[pattern=north east lines, pattern color=black, draw=black] coordinates {
  (TAAD,3.9) (TAAD+GNN,2.6) (TAAD+DST,33.4) (Ablation,11.0)
};
\addlegendentry{No BBox}

\end{axis}
\end{tikzpicture}
\caption{Overall recall and recall by visibility subset. Only TAAD+DST maintains substantial recall when the acting player has no bounding box (33.4\%), highlighting robustness to missing visual localization.}
\label{fig:recall_bars_bbox}
\end{figure}

\paragraph{\textbf{Robustness to visibility and broadcast conditions}} Roughly 80\% of events occur with a bounding box available for the acting player and 20\% without (with 9\% due to replays). Focusing on recall, when a box is present, all methods perform reasonably well, with TAAD+GNN slightly ahead (see Figure \ref{fig:recall_bars_bbox}). The crucial difference appears without boxes: TAAD and \linebreak TAAD+GNN are almost entirely dependent on visual cues and recover very few events when bounding boxes are missing for the acting players (3.9\% and 2.6\% recall), while TAAD+DST maintains 33.4\% recall. Overall, approximately 9 to 10\% of TAAD+DST’s true positives occur without bounding boxes (compared to about 1\% for TAAD). This shows that TAAD+DST can infer actions even when the actor is not visible and highlights the value of tactical priors and long-range consistency. Even when the actor is visually missing, due to occlusion, off-screen positioning or replay cuts, game-state–aware denoising can still infer plausible and temporally consistent actions.

\bigskip

\paragraph{\textbf{Ablation analysis and role of tactical context}} The ablated TAAD+DST, trained without multi-agent reasoning, confirms that structured context is essential. It underperforms the full TAAD+DST across all aggregate metrics (F1: 49.1\% vs. 67.5\%) and in AP for most classes (Fig.~\ref{fig:ap_radar_sorted}). Yet, it exceeds the visual baselines in several cases and, notably, outperforms TAAD and TAAD+GNN in recall and F1 at low confidence thresholds when no bounding boxes are available, particularly for \textit{Throw-in}, \textit{Drive}, \textit{Pass} and \textit{Cross}. This behavior suggests that the ablated version still leverages local spatiotemporal regularities, such as the recurrent spatial patterns of throw-ins near the sidelines and short-term temporal continuity, but lacks the global, role-aware reasoning that enables the full TAAD+DST to capture coordinated team behavior across longer sequences.

The comparison between the ablated and full TAAD+DST therefore shows that temporal modeling alone cannot explain the observed improvements. The full model’s gain arises from integrating tactical priors and multi-agent relationships over longer time horizons. Two mechanisms likely underpin this effect. First, role and team conditioned priors encode who is likely to perform what and where; for example, wingers are more likely to cross, central roles to pass or shoot, and defenders to clear or block. Second, long-range temporal reasoning enforces ball possession continuity and smooths noisy clip-level detections, allowing the model to infer missing or off-screen actions and filter implausible events. Together, these mechanisms yield precision gains at fixed high recall and enable action recovery under occlusion or replay conditions.

\section{Conclusion}
\label{conclusion}
This work presents FOOTPASS, a public benchmark for player-centric action spotting that couples full-match broadcast video with multi-modal, multi-agent tactical context. Using FOOTPASS, we provide reproducible baselines confirming the previously reported benefits of tactical structure and long-range temporal reasoning on broadcast-level data, while quantifying their robustness under occlusion and replay conditions. The benchmark establishes a solid foundation for evaluating methods on realistic soccer footage that reflects the challenges of professional match analysis.

On the dataset side, future work includes extending annotations with acting-player and referee boxes, hierarchical action taxonomies, and sparse events such as infractions, referee interventions, and set pieces. On the methods side, we plan to explore end-to-end learning that unifies perception, tracking, and sequence reasoning; leverage audio and commentary to recover off-screen or occluded actions; and evaluate direct spotting systems that predict $(\text{frame}, \text{team}, \text{jersey}, \text{class})$ from video without explicit tactical inputs.

FOOTPASS thus opens new avenues for research at the intersection of computer vision and tactical modeling, providing a realistic and publicly accessible foundation to study how visual cues and tactical structure, such as spatiotemporal data and roles, can be jointly leveraged for player-centric action understanding.

\bigskip

\paragraph{\textbf{Data availability}} The FOOTPASS annotation files supporting this study will be released on Hugging Face and the baselines on Github\footnote{\url{https://github.com/JeremieOchin/FOOTPASS}}, where the final license and README will specify terms of use for research and evaluation.

The associated broadcast videos are distributed via the SoccerNet dataset repository on Hugging Face and remain subject to SoccerNet's current non-disclosure agreement (NDA). Researchers wishing to access the footage must request it directly from SoccerNet and agree to the NDA; our experiments can be reproduced by combining those videos and annotations with the code provided in the Github release.

Any additional materials required to reproduce the results are available from the corresponding author upon reasonable request.

\paragraph{\textbf{Declaration of generative AI and AI-assisted technologies in the writing process}} During the preparation of this work, the authors used ChatGPT, developed by OpenAI, exclusively in order to improve grammar, spelling, and \LaTeX{} formatting. After using this tool, the authors reviewed and edited the content as needed and take full responsibility for the content of the publication.



\bibliographystyle{elsarticle-harv} 
\bibliography{cas-refs}

\end{document}